# Towards Robust Classification with Image Quality Assessment

Yeli Feng
Nanyang Technological University
yfeng002@e.ntu.edu.sg

Cai Yiyu
Nanyang Technological University
mycai@ntu.edu.sg

**ABSTRACT**

Recent studies have shown that deep convolutional neural networks (DCNN) are vulnerable to adversarial examples and sensitive to perceptual quality as well as the acquisition condition of images. These findings raise a big concern for the adoption of DCNN-based applications for critical tasks. In the literature, various defense strategies have been introduced to increase the robustness of DCNN, including re-training an entire model with benign noise injection, adversarial examples, or adding extra layers. In this paper, we investigate the connection between adversarial manipulation and image quality, subsequently propose a protective mechanism that doesn't require re-training a DCNN. Our method combines image quality assessment with knowledge distillation to detect input images that would trigger a DCCN to produce egregiously wrong results. Using the ResNet model trained on ImageNet as an example, we demonstrate that the detector can effectively identify poor quality and adversarial images.

## I. INTRODUCTION

Following the groundbreaking leap of deep convolution neural network in object classification tasks [1], numerous advances have been made in the network architecture [3,4,5], training and transfer learning techniques [6,7]. Many researchers are motivated to apply DCNN to tackle various visual pattern recognition problems. For example, object detection and localization [8], image and video caption [9], facial recognition [10], and more. With these researches, DCNN-based applications have flourished in recent years and in many areas including medical imaging diagnosis [11], autonomous driving [12], super resolution image [13], aesthetic quality assessment [14] as well as many other industries and consumer sectors.

DCNN's near human-level performance in terms of accuracy naturally raises questions such as why it performs so well; what is precisely learned inside each neural layer; and is it as robust as the human vision when quality degradation is present in the images.

Image quality assessment (IQA) [15] as a research topic has existed since the beginning of the image processing discipline. The types and magnitude of image quality distortion, as well as their impacts on the visual quality perceived by the human visual system have been extensively studied. Numerous IQA algorithms have been developed and have found their ways in many image processing and multimedia applications such as denoising and super-resolution. In very recent years, studies emerge around the impact of image quality on DCNN performance. DCNN trained on the ImageNet suffer non-trivial performance degradation [16] when images are distorted with high degree of blur or noise. Significant performance degradation also occurs in the DCNN-based face recognition model [17] trained on popular face datasets. [18] reports that Google's Cloud Vision API is also vulnerable when fed with noisy images.

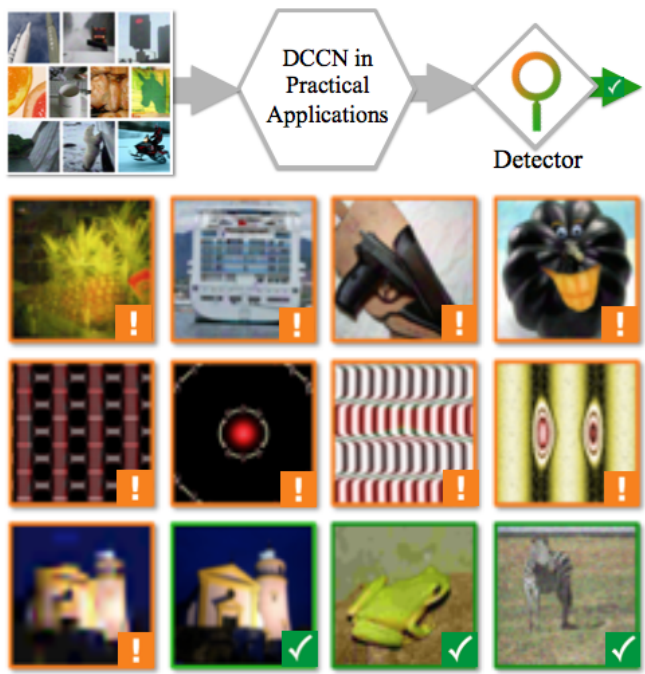

**Fig. 1.** Images with an alert icon are identified as not fit by our detector. Images with a green tick are labeled as fit by our detector. Original images of the top and bottom rows are from the validation set of ImageNet [2]. Images in the second row are adversarials perturbed with the universal perturbation map [21] and classified by ResNet-152 in left to right order as goldfish, vending machine, sock, and bow tie. Images in the third row are examples from [20] and classified by ResNet-152 in left to right order as honeycomb, traffic light, accordion and water snake. The bottom row includes images injected with blur or Gaussian noise at different distortion levels and classified by ResNet-152 in left to right order as whiskey jug, beacon (true label), tree frog and zebra.

The investigation on DCNN reveals its vulnerabilities as well as discovers many methods that can generate adversarial examples to elicit utterly wrong results from DCNN. [19] is the first in the literature to demonstrate that when given a clean image, iteratively exploiting computed gradient and perturbing pixel values in a range that is indistinguishable to the human eye, we can fool DCNN to misclassify a school bus as an ostrich. Using evolutionary algorithms, [20] shows that DCNN can be easily fooled into making highly confident predictions on images that are unrecognizable to human eyes. Very recently, [21] proposes a method that produces a single perturbation map from a trained DCNN model, dubbing the universal perturbation map. By using several popular DCNN networks, the team demonstrates that DCNN is vulnerable to fooling images perturbed with these maps. According to [22], this universal perturbation method is one of the most potent adversarial methods so far in the literature.

Others look into the physical world for potential threats to DCNN-based applications. [23] finds that the ImageNet Inception classifier [24] misclassifies a significant percentage of images obtained from cell-phone cameras. Toxic road signs created by taking a black-box approach to distort road signs with lenticular printing and background injection techniques [25] have also been found to deceive auto-driving cars successfully. By using a forensic task, camera model identification, [26] shows that DCNN is exceptionally vulnerable to adversarial examples generated with Fast Gradient Sign (FGSM) and projected gradient descent (PGD) methods.

Disregarding the generation methods, we observe that a significant percentage of malicious alternations in the adversarial images in the literature are perceivable to the human eyes if seen under a viewing condition that is adopted in the IQA research field to collect subjective quality scores from human observers. Most of these alternations bear some resemblance to signal distortions in the color image dataset [27] used to assess and compare visual quality metrics and algorithms. An extreme example is the unrecognizable fooling images produced using the gradient ascent method (see the bottom row in Appendix C). The visual qualities of these images are not fit for carrying out any object detection task.

Our study shows that the classification performance of DCNN drops when the quality of input images degraded. Image quality is often ignored in visual recognition challenges such as PASCAL VOC [28] and ILSVRC [2], where most if not all images used for model training and validation are recognizable or around average quality. However, in reality, vision applications could receive images far below average quality. For example, auto-driving in an extremely hazy day or image acquisition lens is dirtied. Such scenarios pose practical challenges to the design of robust vision applications.

By gauging adversarial and poor-quality images with a visual quality metric, we propose a method to detect images that are not fit for carrying out an object classification task. One of the uniqueness of the proposed method is that it is non-intrusive per the object classification system. We do not alter the structure of the DCNN and its trained weights. Instead, we conjugate its output (a prediction vector) with an image quality score to train an auxiliary network. This network is capable of detecting input images that are not fit for the object classification task, i.e., the object classification system would produce a wrong result with a high probability if not stopped.

In this paper, related works are reviewed first, followed by a correlation study between the performance of DCNN and image quality and adversarial examples. The proposed detection methods are explained in section IV, followed by experiments and analyses in section V. Finally, conclusion and future work are discussed in section VI.

## II. RELATED WORKS

On protecting CNN against adversarial attacks, distinct approaches have been explored and demonstrated by many researchers. One direction is modifying the images by injecting small noise and minor adversarial perturbation into clean data and then training or re-training a DCNN with both original and processed data. It is also shown that training with compressed images mitigates the adversarial impact to some degree. Another direction is modifying the CNN structure by adding an extra layer to smooth out small adversarial noise or devising a regularization strategy to increase the stability of neuron activations. This line of approach has been shown effective on smaller networks trained with MNIST and CIFAR datasets but computationally hard to re-train very large networks that were trained on the entire ImageNet data. A third direction is attaching an auxiliary network to the intermediate layers or output layer of a DCNN and using these layers' output as features to learn a detector that discriminates adversarial images from clean images. Our work is closer to the third direction in that the proposed method also utilizes auxiliary networks. Related works in this direction are discussed below.

Quantizing the output from the late-stage ReLU layers of VGG-19 and ResNet-32 into discrete input features, [22] trains an RBF-SVM as an adversary detector and reports performance on four types of adversarial attacks including the DeepFool [29]. It shows that adversarial manipulations in the depth map of RGBD images can be detected as well.

Collecting convolutional layers' output from VGG-16 and AlexNet and then processing them with PCA analysis, [30] extracts the statistics from normalized PCA coefficients as an input feature to a sequence of base classifiers which form a cascade SVM. The SVM is trained with 2000 L-BFGS adversarial images generated from the original images of the ImageNet validation set and a subset of the original images. The adversarial detection performance is tested with 2000 L-BFGS adversarial images. The model generalization capability is tested using the fooling images in [20].

Without requiring adversarial images, [31] uses autoencoder to learn the manifold of clean images. Their detector will reject a test image if its reconstruction error is significant or the probability divergence between the target classifier and the encoded representation is big. The authors conclude the detector performs well over MNIST and CIFAR-10 images perturbed by DeepFool.

[32] proposes to extend a smaller subnetwork from ResNet-32 as an adversary detector. The detector itself is a CNN and trained with adversarial images in various degrees of perturbations. Using images from the CIFAR-10 and relevant ten object classes from the ImageNet, the authors conclude the detector has strong discriminative power over tiny perturbations even almost invisible to humans.

[33] proposes a detection framework by measuring the difference between object classification result of a given image and that of its reduced images using reducing color bits and smoothing processing. The detection performance of eleven attacks over six types of adversarial methods is reported over MNIST, CIFAR-10, and a small portion of images from the ImageNet validation set.

[34] proposes a detection strategy that compares the difference between classification results of a given image and that of its reduced images is exploited as well. The reduced images are produced using scalar quantization and smoothing processing. Performance is reported on the MNIST and three object classes from the ImageNet only.

On the impact of image quality to the performance of CNN, [16] investigates the performance degradation on object classification task over two types of distortion Gaussian blur and additive Gaussian noise. In [17], the face recognition task is used to study the decrease in performance over six kinds of distortion. [18] finds Google's Could Vision API produces different classification given the same image but injected with noises of various degrees and conclude denoising the image before feeding it to the API mitigates the problem.

To our best knowledge, there is no work in the literature connecting the image quality factor to the adversarial attack and propose a holistic solution that tackles both at the same time. The main contributions include:

- From an image quality assessment perspective and using semantic distance, we thoroughly study the impacts of four common types of distortion injection and three types of adversarial manipulation to object classification outcomes. We consider the semantic distance is a better metric in the classification context because the information on how wrong the result produced is encoded. Give an example; in Wu-Palmer [35] measure catamaran as 0.92 similar to trimaran and .052 similar to drilling platform.
- Experimentally, we show the signal characteristic shared between distortion injections and adversarial perturbations and propose the image quality as a useful feature component for adversarial detection.
- Using a no-reference IQA model, state-of-art object classification CNN trained on the ImageNet, and experimental data that covers 1000 object classes from the ImageNet, we demonstrate that the proposed method can effectively detect poor quality images and protect the object classification CNN from the adversarial attacks tested in a meaningful percentage as well.

### III. IMAGE QUALITY AND CLASSIFICATION PERFORMANCE

Studies on image quality and adversarial attacks have a different motivation. In IQA, image pixels are altered by software to simulate perceptual quality degradation per human eyes. In adversarial attacks, image pixels are altered algorithmically to trigger DCNN into produce wrong outputs by human standards. Although the methods are different as well, pixel values are altered in both fields. Basing on this fundamental, we theorize that IQA methods could be leveraged into the design of adversarial defense methods. In this section, we describe the investigation methodology and present the experimental findings that support our theory.

*A Methodology*

Traditionally, image quality is systematically studied in types and magnitudes of distortion [27, 36, 37]. Various imaging processing techniques are applied to pristine images to produce distortions. In a laboratory or semi-controlled environment, the distorted images are presented to human observers to evaluate the visual quality, with or without the pristine image aside. The quality scores collected from human observers are consolidated into a subjective mean opinion score (MOS) for each distorted image.

The image quality assessment researches computational models that estimate the quality of an image that is perceived by the human visual system. Based on whether the reference image of a distorted image is available or not, an IQA model is either of full-reference, reduced-reference, or no-reference. To the problem in hand, no-reference IQA is a suitable tool. Most learning-based no-reference IQA models start with a quality-aware feature extractor that is designed manually. Subsequently, the extracted features are fed to a support vector machine to train a quality prediction model. For example, in [38], the quality aware features are general Gaussian distribution (GGD) statistics fitted from the coefficients of a discrete cosine transform from pixel intensity in the spatial domain. In [39], the underlying signals are histograms of pixel intensity, and then they are compressed into a learnable feature using GGD statistics.

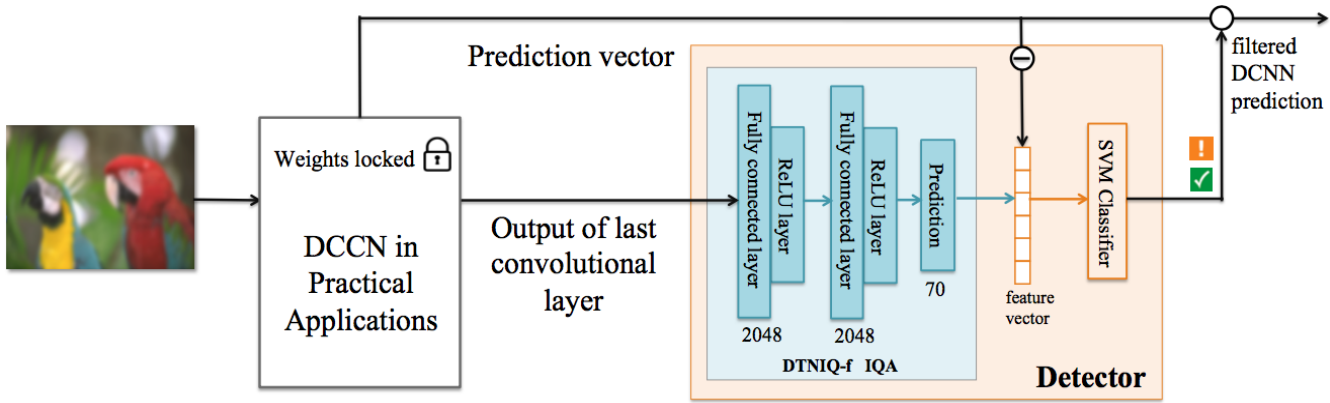

**Fig. 2.** A schematic diagram of the proposed detector

[40] designs a 14-layers CNN-based no-reference IQA model to extract features from 32x32 patches and learn a quality regressor. Arguing that not every region in an image has an equal impact on the overall image quality, a patch-based weight map is also learned to pool the quality predictions into one score. [41] designs a 6-layers CNN-based no-reference IQA model to learn a regressor. [42] proposes a 4-layers CNN architecture. In this study, we use a CNN-based no-reference IQA model DTNIQ-f from [43] to obtain the visual quality scores of distorted and adversarial images.

The DTNIQ-f design abandons the manual feature extraction step, instead uses the convolutional layers of popular CNN object classification models [1,3,5] trained on the ImageNet as a generic image feature extractor and trains a quality predictor with datasets from the IQA domain. As shown in Fig. 2 the DTNIQ-f IQA model contains two fully connected layers of 2048 neural nodes and an output layer of 70 nodes. The activations from the fully connected layers are filtered by Rectified Linear Units (ReLUs). [6] and subsequent researches show that compared to the traditional tanh and sigmoid activation units ReLUs converges faster and leads to a better solution. Predicted MOS score is computed from the prediction layer using softmax. Let $\bar{v}$ be the classification score and $p$ the number of patches cropped from one image during testing, the predicted MOS is calculated using $q = \frac{1}{p}\sum_{i=1}^{p}\sum_{j=1}^{70} j * v_j$ .

For investigation, we collect 4966 distorted or pristine images from the traditional image quality research literature, 5000 unrecognizable fooling images from [20]; and using color images only from the ImageNet validation set and the universal perturbation [21] method produces 49,056 recognizable adversarial images, and using the gradient ascent method produces 3000 unrecognizable adversarial images.

Firstly, the object recognition results of 4966 images from IQA datasets [27, 36, 37] are produced using DCNN models trained on the ImageNet. The semantic distances of results between distorted and reference images are calculated in Wu-Palmar term [35]. Using semantic similarly, we measure the impact of image quality to DCNN. From Table 1, we see AlexNet, VGG-16 and ResNet-152 are all affected. To CSIQ dataset, the mean similarity drops to 0.799. See appendix A for imagery examples about semantic similarity measured in the Wu-Palmar term. Nevertheless, the performance degradation of the three DCCN models is on par. Considering ResNet-152 is more widely used in the recent literature, we use ResNet-152 only for the subsequent analyses and experiments.

The subjective quality scores of [27] are in MOS form ranging from 0 to 9, whereas difference MOS (DMOS) form is used in [36, 37] with ranges from 0 to 1 and from 1 to 100 accordingly. For comparison, the original scores in these datasets are normalized to MOS from 0 to 1 in Table 1. The MOS scores predicted by the IQA model will be normalized to [0,1] also.

TABLE I: IMPACT OF IMAGE QUALITY ON PERFORMANE OF WELLKNOWN DCNNS TRAINED ON IMAGENET

| | | | Median | Mean | Std. |
|---|---|---|---|---|---|
| TID2013 [27] | Subjective MOS | | 0.511 | 0.497 | 0.138 |
| | Wu-Palmar Similarity | AlexNet | 1.000 | 0.825 | 0.260 |
| | | VGG-16 | 1.000 | 0.874 | 0.191 |
| | | ResNet-152 | 1.000 | 0.894 | 0.196 |
| CSIQ [36] | Subjective MOS | | 0.677 | 0.649 | 0.263 |
| | Wu-Palmar Similarity | AlexNet | 1.000 | 0.800 | 0.257 |
| | | VGG-16 | 1.000 | 0.826 | 0.249 |
| | | ResNet-152 | 1.000 | 0.799 | 0.263 |
| LIVE [37] | Subjective MOS | | 0.426 | 0.490 | 0.300 |
| | Wu-Palmar Similarity | AlexNet | 0.800 | 0.767 | 0.251 |
| | | VGG-16 | 1.000 | 0.851 | 0.225 |
| | | ResNet-152 | 1.000 | 0.835 | 0.238 |

## B  Image Distortions

The visual signal in the image and video applications are subject to many kinds of distortions before presented to human or machine viewers. The distortions could be introduced during the signal acquisition, transforming, compression, transmitting, and restoring phases. Inarguably the poorer quality of input image, the less accurate object classification will be [19]. We are curious 1) whether there is a simple linear coloration between the image quality and DCNN performance in terms of accuracy, and 2) given images of similar quality, where there is a type of distortion impacting DCNN performance less.

Analyzing the classification outputs from ResNet-152 on 3000 distorted images from TID2013 by distortion types, several distinct patterns can be observed from Fig. 3. ResNet-152 is reasonably robust to images distorted by masked noise (MN), none eccentricity pattern noise (NEPN), the mean shift of intensity (MS), and a few more. Their kernel density estimation (KDE) plots indicate the semantic distance of these distorted images to corresponding reference image hold close to 0.9 while their visual quality varies from poor to OK quality, i.e., from 0.3 to 0.7 in predicted MOS. On the other hand, DCNN performance is highly sensitive to local block-wise distortion (BLOCK) and change of color saturation (CCS), as their KDEs stretch down to 0.3 along the semantic similarity axis while the image quality varies within a smaller range.

From Fig. 3, we can also make a qualitative observation that the positive correlation between CNN performance and input image quality is not linear, given that the performance degradation of DCNN responds differently to different types of distortions. Remember, these distorted images are generated from the same set of reference images.

## C  Adversarial Images

[19] shows hardly observable perturbation of image pixel values can cause CNN to produce a ridiculously wrong classification result. Several studies have been done to fool CNN. On the other side, methods defending CNN against such potential attacks are emerging. These two research directions are forging an arms race dynamic. Through observing the quality of the fooling images produced with the gradient ascent method in [20], an immediate intuition arises: could image quality be a useful feature to block CNN from being fooled or attacked?

In [43], we lock down the weights of the ResNet-152 that is pre-trained on the ImageNet. We then apply transfer learning to build no-reference IQA models. Using the DTNIQ-f IQA model trained with the TID2013 dataset, we evaluate the quality of the unrecognizable and recognizable adversarial images, in predicted MOS. Two types of unrecognizable adversarial images are assessed. We collect 5000 fooling images from [20] by the evolutionary algorithm and generate another 3000 fooling images using the gradient ascent method. For recognizable adversarial images, we apply the universal perturbation map of ResNet from [21] to perturb original images from the ImageNet validation set. Excluding gray images and failed to perturb, a total of 49,056 recognizable adversarial images are generated. Gray and too small images, e.g. 90 by 90 pixels, are ignored. A few such recognizable adversarial images are shown in the second row of Fig. 1 and appendix C.

TABLE II: QUALITY OF ORIGINAL IMAGENET VALIDATION SET AND ADVERSARIAL EXAMPLES

| Number of Images | Adversarial Method | Predicted MOS | | |
|---|---|---|---|---|
| | | Median | Mean | Std. |
| 49056 | None | 0.429 | 0.454 | 0.106 |
| 49056 | Universal Perturbation [21] | 0.357 | 0.383 | 0.074 |
| 5000 | Evolutionary Algorithm [20] | 0.271 | 0.289 | 0.122 |
| 3000 | Gradient Ascent Method [20] | 0.271 | 0.271 | 0.018 |

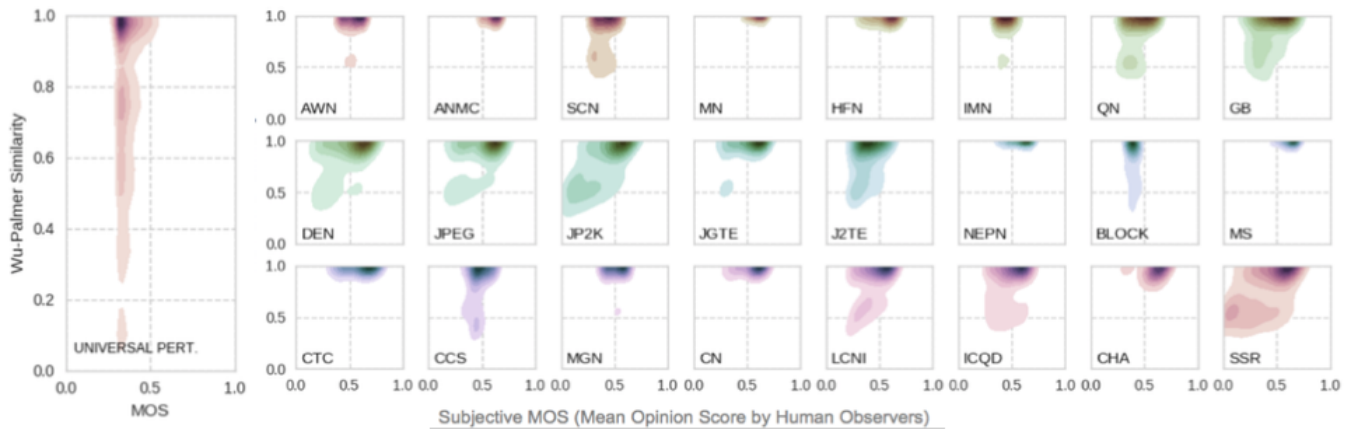

**Fig. 3.** Left is a KDE of Wu-Palmar similarity of adversarial images generated with the universal perturbation map [21] versus quality predicted by the DTNIQ-f IQA model. Right is a grid of KDEs of TID2013 distorted images to their pristine images versus quality in normalized MOS term scored by human observers. The subplot legend is the abbreviation of distortion type. For example, AWN stands for additive Gaussian white noise, NEPN stands for none eccentricity pattern noise. See the full terms of the abbreviations in appendix B.

The overall quality of these adversarial images is significantly poor, as shown in Table II, statistically lower than the three image quality datasets, as shown in Table I. The average quality of perturbed images is statistically lower than their original versions as well. From the left column of Fig. 3, we see the semantic similarly of objects recognized from clean and perturbed images drops close to zero.

To this end, we propose to take into consideration of image quality as a useful feature component in seeking methods that defense CNN from performance degradation and mitigate the probability of being fooled or attacked.

## IV. PROPOSED METHODS

When visibility is low, a human driver will instinctively brake or slow down under normal circumstances. It is logical to extend this behavioral pattern to machine vision systems for their own safety and their human users. The experimental data and analyses around DCNN performance and input image quality lead us to a hypothesis that visual quality is a useful feature to be exploited for DCNN-based applications to detect potential input signal alterations that are not fit for recognition tasks, disregarding the intention behind the distortion of the images.

In supervised learning, a DCNN-based vision model is trained and validated with a cohort of meaningful images. Given images utterly unrecognizable to human eyes, there is no value for a machine vision system to strive to recognize an object from it. From the statistics in Table II, a simple threshold-based baseline method can be constructed. For the fooling images generated from a mean map, the mean and standard deviation is 0.271 and 0.018 in predicted MOS respectively. If setting a quality threshold at 0.3 in predicted MOS, 94.6% fooling images can be discriminated, if the distribution of the visual quality scores is a perfectly normal distribution.

Applying this threshold on to the ResNet-152 model trained on the ImageNet, testing on the validation set shows that 2.44% images are below the threshold 0.3 and will be labeled as not fit for the object classification task, including tiny images of 90 by 90 pixels. A couple of such images are shown in Fig. 4.

However, when the visual quality of input images is better than extremely poor or objects inside are recognizable to human eyes, say around 0.429 in predicted MOS, the fixed threshold-based detection method will not work well. Because the statistics in Table II tells us that roughly 50% of the original images in the ImageNet validation set will be labeled as not fit.

Studying the statistics of prediction vectors produced by the ResNet-152, significant statistical differences can be observed between the original and adversarial images. The median value of the top-1 class confidence score of original images is as twice as high of corresponding adversarial images, whereas its standard deviation is smaller, see table III.

Distilling knowledge from the prediction vector is first proposed by [44], in which prediction probabilities from an ensemble of neural network models are distilled into a single model, and the performance of this single model is improved significantly. Our approach is to combine the quality score from the IQA model with the classification probability score from the ResNet-152, aiming to enhance detection power. The prediction probability vector from ResNet-152 is in 1000 dimensions, but most are close to zero probability residual, thereby minimal knowledge to distill. We take the probability values of the top-5 predictions of an image together with its predicted quality score to form an input feature of six dimensions. Using the -t-SNE technique, we visualize these input features in the left column of Fig. 4. The gray dots represent features learned from the original images in the ImageNet validation ser. The red dots represent features learned from the fooling images produced by the evolutionary method in [20]. We can observe that most of the red dots are well separated from the gray dots, which infers it is hopeful to use SVM learning a useful detector.

TABLE III: STATISTICS OF RESNET-152 PREDICTION VECTORS

| Category of Images | Top-1 Class Confidence Score | | |
|---|---|---|---|
| | Median | Mean | Std. |
| Evolutionary Algorithm [20] | 0.465 | 0.507 | 0.261 |
| Universal Perturbation [21] | 0.452 | 0.508 | 0.292 |
| Original ImageNet Val. Set | 0.929 | 0.800 | 0.244 |

The proposed method can easily be extended to object detection and localization segmentation tasks where the input features to the detector will be extracted from the candidate proposal regions instead of a whole image.

## V. EXPERIMENTS AND ANALYSES

### A Data

A set of experiments is designed to measure how effective the proposed approach will accept input images that ResNet-152 will correctly recognize objects in it and label images that will trigger ResNet-152 to make mistakes as unfit. Four categories of experiment data are collected or generated:

Original images: The top-1 and top-5 classification accuracy of the entire ImageNet validation set is 75.92% and 92.83%, respectively, by the ResNet-152 model we use. Only the original images that are correctly classified per the top-1 prediction are selected, which leaves 37,962 original images in total.
- Distorted images: four types of distortion in five levels are applied to the above original images. The distortion types include Gaussian blur, white noise, and JPEG and JPEG2000 compression noise. A total of 759,240 distorted images are generated. Their quality ranges from 0.071 to 0.729 in predicted MOS.
- Recognizable adversarial images: 37,962 adversarial images are generated from the original images, using the universal perturbation map of ResNet-152 from [21]. The adversarial target of each sample is chosen randomly from the 999 classes, excluding the true object class.
- Unrecognizable adversarial images: 3000 images are generated from a mean map using the ascent gradient method. We denote this group as Fool_GD. And 5000 images are collected from [20] that are produced using the evolutionary algorithm. We denote this group as Fool_EV.

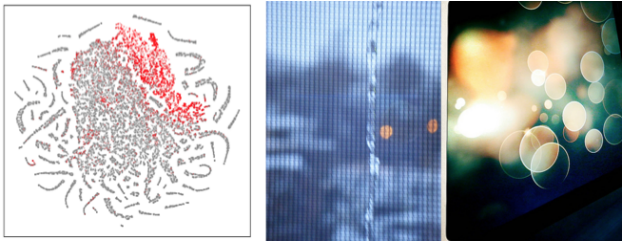

**Fig. 4.** The left column is a t-SNE visualization of input features for the fit-for-task detector. Middle and right columns are images from the ImageNet validation set. The predicted MOS values of both images are below 0.3.

*B   Evaluation*

We use ten DTNIQ-f NR-IQA models trained with the TID2013 image quality dataset to predict MOS scores of distorted and adversarial images. For distorted images, one model is selected randomly from the ten DTNIQ-f NR-IQA models to produce MOS scores, i.e., the ten detectors are trained with a single set of features. Performances are reported in an average of the ten detectors. For recognizable distorted and adversarial images, a baseline of object recognition results from ResNet-152 is established first. The detection rate and the recognition accuracy of images labeled as fit by the detector are compared with the corresponding baseline.

All experiments and data generation are carried out on a computer equipped with a GeForce GTX 1080 Ti and using the Caffe platform, with exception to recognizable adversarial images that are generated on the TensorFlow platform.

*C   Baselines*

Not all attacks are successful, and not all distortions lead to incorrect object recognition outcomes. We define object recognition accuracy of ResNet-152 without quality awareness as baseline-1 and that with the threshold-based quality filtering as baseline-2. Table IV summarizes the test results. In baseline-1, we see the accuracy of ResNet-152 on adversarial and distorted image groups drops from 100% to just above 50%. Note the corresponding original images that are used to produce these images are all correctly classified by ResNet-152.

In baseline-2, 96.7% Fool_GD and 54.11% Fool_EV images are detected. Since these are fooling images, there is no point to reporting ResNet-152's accuracy. 48.16% of distorted images are detected, which leads to the accuracy of ResNet-152 over fit images increases to 73.04%.

Baseline-2 detects only 4.21% of the recognizable adversarial images. If increasing the quality threshold from predicted MOS 0.3 to 0.35, 47.25% of them can be detected. However, the accuracy of ResNet-152 over fit images doesn't increase; instead, it drops from 73.04% to 72.45%. This indicates that the discriminative power of quality feature alone is weak over the recognizable adversarial images.

*D   Performance*

Next, we show that by distilling object classification information from ResNet-152, the detection power increases significantly. With input features generated from four groups of images, many different training schemas are possible. However, our primary goal of this experiment is to improve the detection capacity over the recognizable adversarial images. After some trials, we decide to take only the recognizable adversarial images and their corresponding original images to learn a binary detection classifier. 50% of recognizable adversarial images and their corresponding original images are selected randomly to form a training set, which ensures an adversarial and its original image appear together in either the train or test set. Rest 50% is held out for testing. This sampling method is repeated ten times to train ten detection classifiers. To test the detection performance on distorted images, only distorted images that their corresponding original images are not in a training set are selected.

Table V summarizes the test results. We can see 65.95% of recognizable adversarial images are detected, and the accuracy of ResNet-152 over fit images increases to 85.22%. The result indicates that, indeed, there is useful information residing in the prediction probability vector of the object recognition classifier. The detector can effectively protect ResNet-152 and alike-based applications from potential attacks that exploit the universal map adversarial method.

Can the detector generalized well to quality degradation produced with other methods? The detector is not trained with any sample from the fool image groups and distorted image group. The patterns of artifacts injected into the distorted images are different from the universal perturbation map. Nevertheless, the experiment result indicates that the detector can generalize what learned from the latter pattern to discriminate common distortion patterns such as white noise and JPEG compression error. As we can see, the accuracy of ResNet-152 over fit images increases to 92.36%, from 73.04% in baseline-2. For the Fool_EV group, the detection rate increases to 78.29%, from 54.11% in baseline-2. And only 0.06% of images from the Fool_FD group is missed out by the detector.

TABLE IV: SUMMARY OF BASELINES

| Image Group | Baseline-1 ResNet-152 Accuracy | Baseline-2 Detection Rate | | Baseline-2 ResNet-152 Accuracy |
|---|---|---|---|---|
| Adversarial | 55.42% | 4.21% | Fit Images → | 54.69% |
| Distorted | 57.04% | 48.16% | | 73.04% |
| Fool_EV | - | 54.11% | | - |
| Fool_GD | - | 96.70% | | - |

TABLE V: PERFORMANCE OF DETECTOR

| Image Group | Detection Rate | | ResNet-152 Accuracy |
|---|---|---|---|
| Adversarial | 65.95% | Fit Images → | 85.22% |
| Distorted | 65.81% | | 92.36% |
| Fool_EV | 78.29% | | - |
| Fool_GD | 99.64% | | - |

Will the detector trigger a high false alarm rate over images that would be correctly classified by ResNet-152? Tests with the original image group show that the detector labels 91.43% of them as fit. Image examples that the detector succeeds or fails to detect are shown in appendix C.

*E   Comparison and Discussion*

The defense against adversarial work in the literature looks into adversarial examples only. Therefore, most of these works validate their methods against adversarial examples but don't test the false alarm rate over clean images. For example, the cascade SVM in [30] achieves overall 97.34% detection accuracy over the Fool_EV images; however, the potential misclassification rate is not investigated. Using 1000 MNIST samples and 100 random samples from the ImageNet, [34] concludes the cascade SVM produces a 92% false positive rate.

Most of these works use images from the MNIST and CIFAIR-10 to validate their defense methods. If images from the ImageNet are involved, only a small portion is selected. For example, [33] tests 5000 images from ten object classes in the ImageNet that are relevant per CIFAIR-10. [34] validates 1489 ImageNet images belong to three object classes (zebra, panda, and cab), why selecting these particular three object classes is not explained.

We validate proposed methods over 805,202 adversarial or distorted images, and 37,962 original images from the ImageNet validation set, covering 1000 object classes in the ImageNet. It is hard to run a direct performance comparison with other defense methods at this massive scale. Besides adversarial images yet perceivable to human eyes per IQA standard, we are interested in detecting poor quality images too. Alternatively, we construct two baselines and compare the performance of the proposed method with the baselines. And show our detector can well protect the object classification CNN from making wrong predictions when input images are of poor quality or maliciously altered with the three attacking methods tested.

Examining the unrecognizable adversarial images, we can observe these images do not have a meaningful composition from photography standard or distinctly different composition compared to natural images. The composition of an image impacts the visual distinctness of objects in an image, including the space and size of the foreground object, the complexity of background texture, contrast, and many more.

Analyzing the classification results of reference and distorted images in the image quality datasets, we notice a couple of images stand out, see columns 1 and 2 in Fig. 5. ResNet-152 classifies all distorted images of each as the same object class as its reference image. At this time, we don't see a simple way to exploit it, but it could be a potential direction to look into in the future.

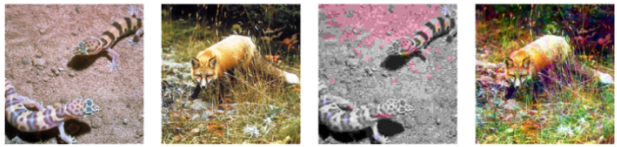

**Fig. 5.** Columns 1 and 2 are reference images that Wu-Palmar similarity of all their distorted images equals 1. Column 3 is a sample of jpeg distortion of normalized MOS 0.451 and column 4 is a sample of additive pink Gaussian distortion of normalized MOS 0.449.

## VI   CONCLUSION

Although poor quality images are not malicious, they could drag the object classification performance down and pose potential harm the same way as adversarial images. In this paper, we identify the value from the traditional IQA field in tackling one emerging challenge to the rising deep CNN technologies. Experimentally we show that there is an intrinsic connection between the adversarial pixel perturbations and visual quality per the human visual system. We introduce the fit-for-task notation and propose non-intrusive methods to prevent the production of object classification results when the inputs are far below optimal, thereby increasing the robustness of CNN-based vision systems in the real-world environment. Visual quality manipulation is relatively cheap to implement, but it could also bring down the performance of a CNN network. We conclude that the proposed methods could serve well as a first-line security mechanism for CNN-based vision applications.

APPENDIX A – SEMANTIC SIMILARITY BETWEEN OBJECTS

Examples of objects recognized from reference and distorted images by the ResNet-152 model trained on the ImageNet and Wu-Palmar similarity between objects, which is range from 0 to 1. Each object is represented by a synset in the WordNet [46]. The Wu-Palmar similarity between warplane and airliner is 0.88, and 0.50 between catamaran and safety pin. The images are from the LIVE [37] and TID2013 [27] datasets.

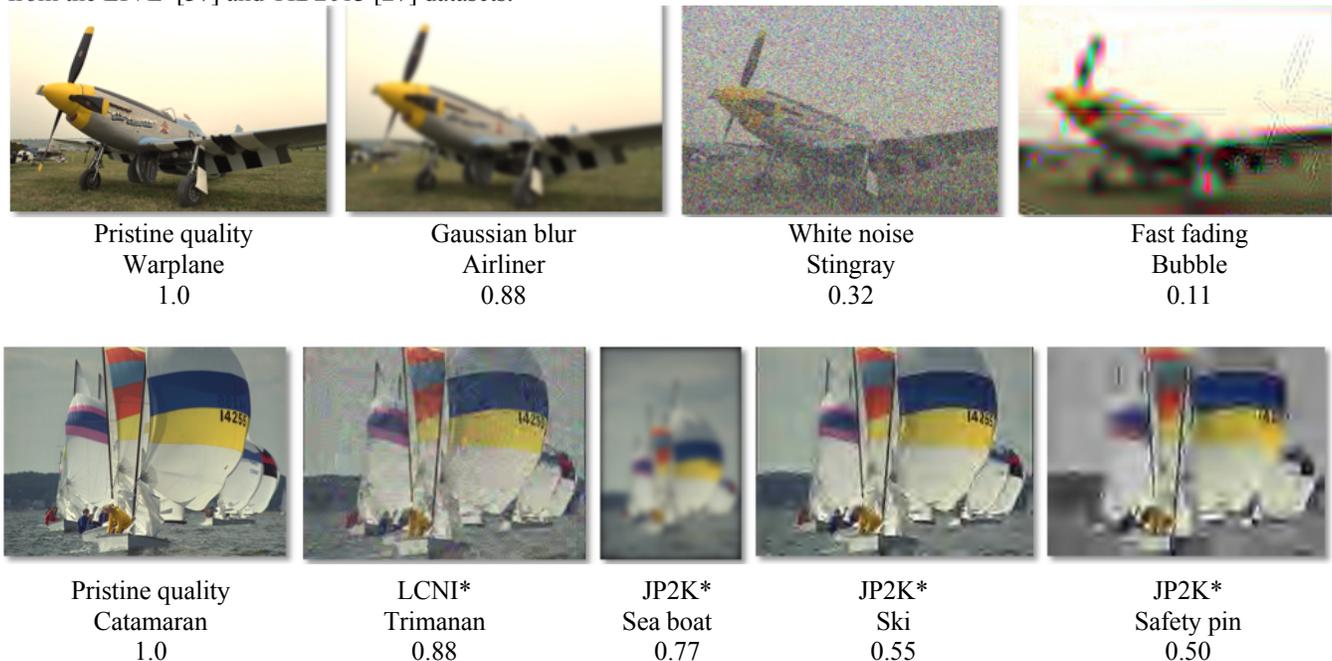

| Pristine quality | Gaussian blur | White noise | Fast fading |
| Warplane | Airliner | Stingray | Bubble |
| 1.0 | 0.88 | 0.32 | 0.11 |

| Pristine quality | LCNI* | JP2K* | JP2K* | JP2K* |
| Catamaran | Trimanan | Sea boat | Ski | Safety pin |
| 1.0 | 0.88 | 0.77 | 0.55 | 0.50 |

* See appendix B for full term of distortion type.

APPENDIX B - FULL TERM OF DISTORTION TYPES IN FIG. 3

| Abbreviation | Distortion Type | Abbreviation | Distortion Type |
|---|---|---|---|
| AWN | Additive Gaussian white noise | JP2K | JPEG2000 compression |
| SCN | Spatially correlated noise | JGTE | JPEG transmission errors |
| MH | Masked noise | J2TE | JPEG2000 transmission errors |
| HFN | High frequency noise | NEPN | Non eccentricity pattern noise |
| IMN | Impulse noise | MS | Mean shift (intensity shift) |
| QN | Quantization noise | CTC | Contrast change |
| GB | Gaussian blur | CCS | Change of color saturation |
| DEN | Image denoising | MGN | Multiplicative Gaussian noise |
| JPEG | JPEG compression | CN | Comfort noise |
| ANMC | Additive noise in color components (more intensive than additive noise in the luminance component) | LCNI | Lossy compression of noisy images |
| | | BLOCK | Local block-wise distortions of different intensity |

APPENDIX C – DETECTION OF ADVERSARIAL EXAMPLES

Adversarial examples in rows 1 and 2 are successfully detected otherwise would be wrongly classified by ResNet-152. In row 3, the left two adversarial examples are falsely labeled as unfit by the detector, although ResNet-152 would correctly classify them; the right two adversarial examples are not detected and successfully fool ResNet-152. The actual object labels of the 3rd and 4th images are goldfish and banana. But ResNet-152 classifies them as brain coral and pineapple. All fooling images in row 4 are detected successfully.

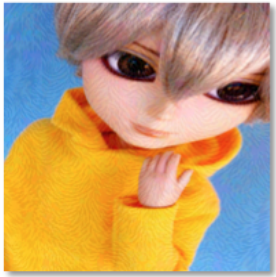 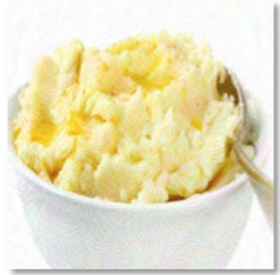 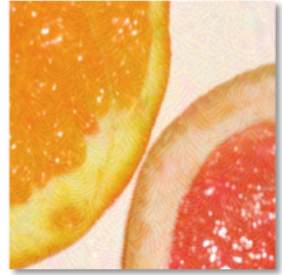 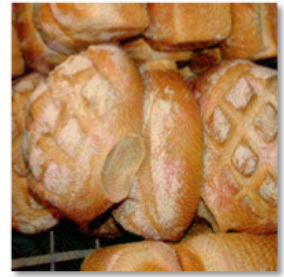

Detected MOS 0.700    Detected MOS 0.614    Detected MOS 0.600    Detected MOS 0.586

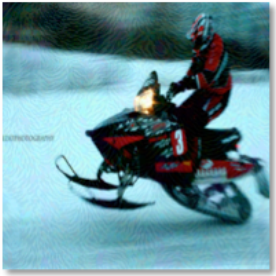 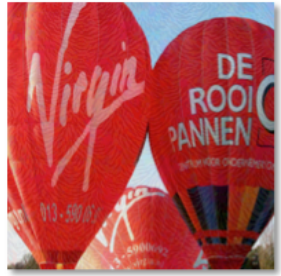 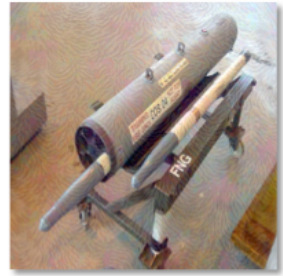 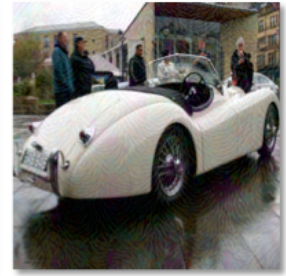

Detected MOS 0.486    Detected MOS 0.471    Detected MOS 0.457    Detected MOS 0.414

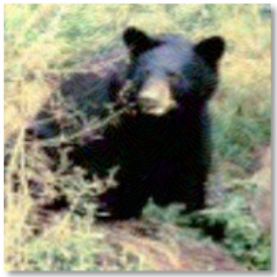 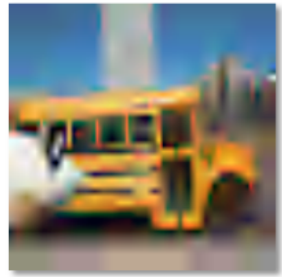 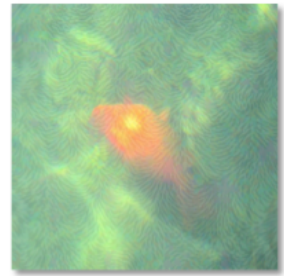 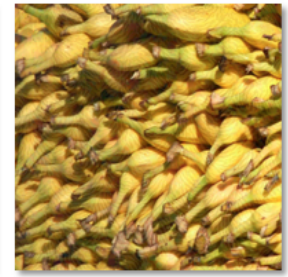

False alarm MOS 0.271    False alarm MOS 0.129    Missed out MOS 0.686    Missed out MOS 0.286

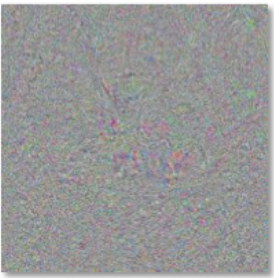 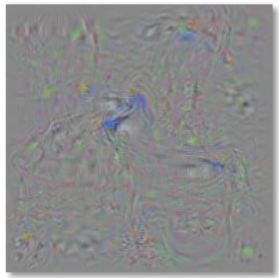 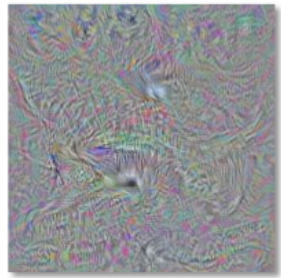 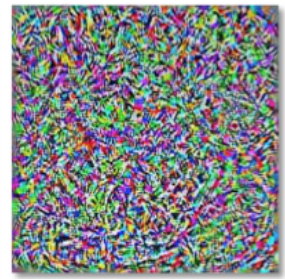

Detected MOS 0.271    Detected MOS 0.271    Detected MOS 0.271    Detected MOS 0.229